# AI-Powered Predictions for Electricity Load in Prosumer Communities[1]

**Aleksei KYCHKIN[2], Georgios CHASPARIS[3]**

[2] Softwarepark 32a, A-4232 Hagenberg, Tel: +43 50 343 813, Email: aleksei.kychkin@scch.at, Web: http://www.scch.at

[3] Softwarepark 32a, A-4232 Hagenberg, Tel: +43 50 343 857, Email: georgios.chasparis@scch.at, Web: http://www.scch.at

**Abstract:** The flexibility in electricity consumption and production in communities of residential buildings, including those with renewable energy sources and energy storage (a.k.a., prosumers), can effectively be utilized through the advancement of short-term demand response mechanisms. It is known that flexibility can further be increased if demand response is performed at the level of communities of prosumers, since aggregated groups can better coordinate electricity consumption. However, the effectiveness of such short-term optimization is highly dependent on the accuracy of electricity load forecasts both for each building as well as for the whole community. Structural variations in the electricity load profile can be associated with different exogenous factors, such as weather conditions, calendar information and day of the week, as well as user behavior. In this paper, we review a wide range of electricity load forecasting techniques, that can provide significant assistance in optimizing load consumption in prosumer communities. We present and test artificial intelligence (AI) powered short-term load forecasting methodologies that operate with black-box time series models, such as Facebook's Prophet and Long Short-term Memory (LSTM) models; season-based SARIMA and smoothing Holt-Winters models; and empirical regression-based models that utilize domain knowledge. The integration of weather forecasts into data-driven time series forecasts is also tested. Results show that the combination of persistent and regression terms (adapted to the load forecasting task) achieves the best forecast accuracy.

## 1 Introduction

The current state of the art in Smart Building technologies provides several opportunities for efficient use of electricity. This can be attained through different forms of demand response, where residential buildings can properly schedule their flexible loads to reduce electricity costs. However, such demand response mechanisms usually require accurate forecasts of the electricity load consumption either at individual buildings or at communities of buildings.

---

[1] This contribution was made possible through the research project Serve-U, FFG #881164. *The research reported in this paper has been partly funded by the Federal Ministry for Climate Action, Environment, Energy, Mobility, Innovation and Technology (BMK), the Federal Ministry for Digital and Economic Affairs (BMDW), and the State of Upper Austria in the frame of SCCH, a center in the COMET - Competence Centers for Excellent Technologies Program managed by Austrian Research Promotion Agency FFG.*

However, generating accurate day-ahead forecasts for electricity loads in residential buildings is a rather challenging task due to the stochastic behavior of the users, as well as the variability in the electrical equipment used in each building. Using black-box machine learning models, including recently developed deep-learning architectures may not necessarily provide accurate forecasts in all cases, while the computational and configuration complexity of such models may be prohibitive. To this end, this paper provides a comparative analysis of several variations of deep-learning-based (AI-powered) methodologies and simpler modeling techniques that better incorporate domain expertise as recently introduced by the authors in [1,2].

The proposed methodologies for day-ahead electricity load forecasting are building upon recent articles of the authors [1,2]. They are extended here by a) providing a comparative analysis of several variations of modern deep-learning-based machine learning models, and b) investigating the impact of weather forecasts on improving electricity load forecasts.

## 2 Methodological Approach

Measurements of the electricity load over a period of several months are sufficient to establish reliable day-ahead forecast models, where forecasts of the electricity load are provided over the following day (i.e., 24 hours as a sequence of 96 samples with 15-min granularity over the following day). Measurements were collected from three residential buildings in the state of Upper Austria. In addition, an artificial community has been established by also considering the aggregated sum of the electricity load consumption as a predicted target variable. In this work, we provide a comparative analysis of deep-learning-based (AI-powered) forecasting models with a collection of simpler forecasting models, namely persistence models, auto-regressive based models, and their combinations.

*Facebook`s Prophet* model [3,4] and *LSTM* model [5,6] constitute modern deep-learning architectures that can be used to establish predictions based on average load consumption through sampling of similar sequences in the past. In a way, this resembles the persistence models *N-days* and *N-same-days* introduced in [1,2], where the predicted load for a future time interval is generated based on the load at similar time intervals in previous days. Alternative machine learning models that can be used for forecasting also includes the *Holt-Winters* (*HW*) and *SARIMA* models that provide multi-step forecasting via smoothing, season, and trend decomposition. Furthermore, the persistence-based regression models, namely the persistence-based autoregressive model (*PAR*) and the seasonal persistence-based regressive models (*SPR*, *SPNN*) have been introduced by the authors in [1] and combine persistence factors with auto-regressive and domain-specific features. In *SPR* and *SPNN* models, we expanded the set of features to capture phenomena that are specifically relevant to electricity load consumption in residential buildings (such as, maximum energy consumption over one day). Furthermore, using *PAR* as a basis model, we also introduced additional features capturing the weather conditions (which are also provided as forecasts), namely the *solar radiation* and *outdoor temperature,* which formulated the *PAR-W* model. A more detailed description of the short-term load forecasting models used is provided below.

## 2.1 Persistence Models

Persistence models constitute computationally efficient forecasting models that can be used for baseline load predictions. In many cases it is useful to know whether the developed forecasting model can give better forecasts than such simple baseline models. The paper considers persistence model using *N* last days of the same day of the week. For example, if it is necessary to forecast the load for the next day, and this day is Monday, then the forecast will be equal to the average load at the same time interval for the previous *N* Mondays. The abbreviation *N-same-days* is used to denote these models. The formal definition will be as follows. Let $y_d(t)$ be the load at time $t$ on day $d$. The persistence model (PM) considering $N$ previous weekdays assumes that:

$$\hat{y}_d^{PM}(t) = \frac{1}{N} \sum_{i=d-7N}^{d-7} y_i(t).$$

Since the electricity load is highly dependent on the users' actions (i.e., their energy use schedules) in the recent past, it is possible to further improve the persistence model by considering a sequence of $N$ days. This model is denoted as the *N-days* model. According to this model, the forecast is generated as follows:

$$\hat{y}_d^{PM}(t) = \frac{1}{N} \sum_{i=d-N}^{d-1} y_i(t).$$

## 2.2 Machine Learning Models for Smoothing, Season, and Trend Decomposition

### 2.2.1 Triple Exponential Smoothing on Holt-Winters Model

The *Holt-Winters* model is designed to identify the trend and seasonality in time series data. The seasonal component of the model explains the recurring fluctuations and is described by the season length, i.e., the period after which the fluctuations begin to repeat. For each observation during the season, its own seasonal component is formed.

In particular, the estimate of the future value of $y$ at time $t$ according to the Holt-Winters model is formed as follows:

$$\hat{y}_d^{HW}(t) = L(t-k) + kP(t-k) + S(t-T)$$

where $L(t)$ is the level component, which is given by:

$$L(t) = \alpha(y(t) - S(t-T)) + (1-\alpha)\big(L(t-1) + P(t-1)\big)$$

$P(t)$ is the trend component, which is given by:

$$P(t) = \beta(L(t) - L(t-1)) + (1-\beta)P(t-1)$$

$S(t)$ is the seasonal component, which is given by the formula:

$$S(t) = \gamma(y(t) - L(t)) + (1-\gamma)S(t-T)$$

where $k$ is the prediction range $k = 96$, $y(t)$ is the actual value of electricity load at time $t$, $T$ is the time series period, $\alpha$ is the data smoothing factor, $\beta$ is the trend smoothing factor, and $\gamma$ is the seasonal smoothing factor. Also, $\alpha, \beta, \gamma \in (0,1)$. In the case of sequential forecasting, $y(t)$ is replaced by the corresponding forecast at the same time.

### 2.2.2 Seasonal Autoregressive Integrated Moving Average Model (SARIMA)

As a variation of the ARIMA model, the SARIMA model, can be used to track the seasonal component of a time series. In this model, parameters $(p, d, q)$ are considered as non-seasonal parameters and remain the same as in the previous model. These three parameters together account for seasonality, trend, and noise in the data sets. Specifically,

- $p$ is the order of autoregression, which allows previous values of the time series to be considered,
- $d$ is the order of integration, which allows previous differences of the time series to be considered,
- $q$ is the moving average order, which allows the model error to be specified as a linear combination of previously observed error values.

In addition to these parameters, parameters $(P, D, Q)$ are applied to the seasonal component of the time series. In addition, a parameter $S$ is added to describe the season length of the time series (that is $96$ if the season corresponds to one day, $7 \times 96$ if the season corresponds to one week, etc., with $96$ being the number of sensor measurements taken within one day). Like the Holt-Winters model, the SARIMA model is a machine learning model that takes into account the seasonality of the series, indicated by $S$. The parameters $(p, d, q)(P, D, Q)S$ were selected by using the Akaike Information Criterion (AIC) and finally fitted as: $(1,1,1)(1,1,1)96$.

## 2.3 Persistence-based Regression Models

### 2.3.1 Persistence-based Autoregressive Model (*PAR*)

Persistence models can detect low-frequency temporal dependencies in the load profile (dependencies occurring over several days or weeks), while autoregressive models can capture high-frequency temporal dependencies (occurring over one calendar day). In references [1,2], we introduced the Persistence-based Autoregressive Model (PAR) that tries to capture the optimal combination of these two types of models. In PAR, the estimated load is defined as

$$\hat{y}_d^{PAR}(t|a_1, \ldots, a_n, b_0) = a_1 \hat{y}_d^{AR}(t-1) + \cdots + a_1 \hat{y}_d^{AR}(t-n) + b_0 \hat{y}_d^{PM}(t)$$

In this case, it is necessary to calculate the set of weights $a_1, a_2, \ldots, a_n, b_0$, corresponding to the optimal combination of high-frequency temporal dependencies (captured by the autoregressive terms) and low-frequency temporal or seasonal dependencies (captured by the persistence terms).

### 2.3.2 Persistence-based Autoregressive Model with Weather Data (*PAR-W*)

An extension of the PAR model also incorporates the weather forecasts. The model retains the benefits of PAR, namely the ability to capture low-frequency changes in the load (i.e., dependencies occurring over several days or weeks) and high-frequency changes (i.e., occurring over one calendar day). It also captures the influence of weather, in particular the solar radiation, to the electricity consumption. The new model, briefly PAR-W, can be written as follows:

$$\hat{y}_d^{PARW}(t) = a_1 \hat{y}_d^{AR}(t-1) + a_2 \hat{y}_d^{AR}(t-2) + \cdots + a_j \hat{y}_d^{AR}(t-j) + b_0 \hat{y}_d^{PM}(t) + c_0 \hat{s}_d(t)$$

where $\hat{s}_d(t)$ denotes the estimate of solar radiation of day $d$ at time interval $t$.

### 2.3.3 Seasonal Persistence-based Regressive Models (*SPR* and *SPNN*)

Seasonal persistence-based regressive models provide higher robustness and ability to exploit causal effects specific to user-behavior changes in user schedules, e.g., users usually consume about the same energy every morning or in certain time periods. Instead of generating the forecast based on the load at the same point in time in the previous day, it is possible to use an average load over an extended window of time in the previous day (the window can be one hour, for example). In addition, the total energy consumption can be another factor that can reduce the uncertainty of the forecasts. A set of features have been introduced to reduce the uncertainty associated with small variations in user schedules. This set of features are defined based on the load recorded on the previous day and on the corresponding day one week ago. In particular, the SPR model is defined as follows:

$$\hat{y}_d^{SPR}(t|a_0, a_1, \ldots, a_{14})$$
$$= a_0 f_d + a_1 y_{d-1}(t) + a_2 y_{d-7}(t) + a_3 y_{rs,d-1}(t) + a_4 y_{rs,d-7}(t) + a_5 y_{h,d-1}(t) + a_6 y_{h,d-7}(t)$$
$$+ a_7 y_{d,d-1}(t) + a_8 y_{d,d-7}(t) + a_9 y_{dh,d-1}(t) + a_{10} y_{dh,d-7}(t) + a_{11} y_{low,d-1}(t) + a_{12} y_{low,d-7}(t)$$
$$+ a_{13} y_{high,d-1}(t) + a_{14} y_{high,d-7}(t)$$

where the features are as follows: $f_d$ corresponds to the type of the day (working day or weekend); $y_d$ is the electricity consumption at day $d$; $y_{rs}$ is the rolling-sum of the electricity load; $y_h$ is the total electricity load within the last hours; $y_{dh}$ is the difference in hourly load; $y_{low}, y_{high}$ corresponds to the low and high energy consumption flag, respectively.

We also explored the possibility of nonlinear dependencies between the above features, by introducing a Multilayer Perceptron (MLP) with the same set of features. The resulting model is called *Seasonal Persistence-based Neural Network* (SPNN) model. This architecture allows for finding non-linear dependencies between the features, which may reveal new patterns in users' behavior.

## 2.4 Generalized Additive Model of Facebook's Prophet

A detailed description of the methodology implemented in the Prophet model of Facebook can be found in references [3,4]. This methodology is based on the Human-in-the-Loop modelling and provide fitting additive regression models (Generalized Linear Models and their extension on Generalized Additive Models, GAM) of the following form:

$$\hat{y}_d^{Prophet}(t) = g(t) + s(t) + h(t) + w_t$$

where $g(t)$ approximates the trend of the series; $s(t)$ captures seasonal fluctuations (daily, weekly, etc.); $h(t)$ captures the effects of holidays and other significant calendar events; $w_t$ is a normally distributed random disturbance. The following methods are used to approximate these functions:

- *trend:* piecewise linear regression or piecewise logistic growth curve,

- *annual or weekly seasonality:* partial sums of a Fourier series with $P$ as a regular period:

$$s(t) = \sum_{n=1}^{N} \left( a_n \cos\left(\frac{2\pi n t}{P}\right) + b_n \sin\left(\frac{2\pi n t}{P}\right) \right)$$

- *"holidays"* (e.g., official holidays and weekends - New Year, Christmas, etc., as well as other days during which the properties of the time series may change significantly - sports or cultural events, natural phenomena, etc.): represented as indicator variables.

Estimation of the fitted model parameters is performed using the principles of Bayesian statistics. The STAN probabilistic programming platform is used for this purpose.

## 2.5 Long Short-Term Memory Model (*LSTM*)

Analysis of the electricity consumption in residential buildings have shown that the statistical characteristics of its weekly variations are almost constant. However, there are huge variations within hourly intervals due to the stochastic nature of users' behavior. In these cases, an in-depth analysis of intra-daily sequences is required, which can be provided by the LSTM artificial neural network due to its ability to learn long-term correlations, efficiently process time series data in short-term sequences, automatically detect and learn patterns of complex sequences and adapt to changing input data [7].

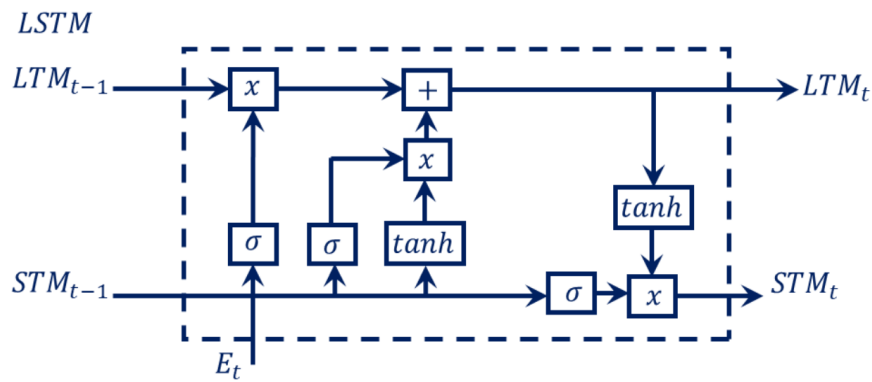

*Figure 1 LSTM architecture with three main components: "forget gate", "input gate" and "output gate"*

It is known that in the LSTM network architecture, the output layer data is first sorted by using the sigma function. The non-stationary source data is excluded, and the remaining data is moved to the next step. The data is sorted using the following expression:

$f_t = \sigma(W_f[h_{t-1}, \alpha_t] + b_f)$,

where $W_f$ is the input weighting coefficient; $h_t$ is the output vector; and $b_f$ is the input layer threshold.

The next process trains the selected input data to determine the prediction indicator and determines their acceptable values. The sigma and tangent functions are used in this process:

$i_t = \sigma(W_f[h_{t-1}, x_t] + b_i)$,

$C_t = tan(W_c[h_{t-1}, x_t] + b_c)$,

where $x_t$ is the input data; $b_i$ and $b_c$ are the neuron thresholds.

Based on the selected data, the neurons in the output layer are identified by systematically combining sigma and tangent functions. That is, the prediction result is determined by the following process:

$$\begin{cases} O_t = \sigma(W_o[h_{t-1}, x_t] + b_0) \\ \quad h_t = O_t \tan(C_t) \end{cases}$$

where $O_t$ is the initial data selected using the sigma function; and $h_t$ is the neuron network output layer.

In this study, we used the following parameters of the LSTM architecture:

- The number of input layer neurons is $96 \times 10$;
- The number of neurons of the first hidden recursive layer is $96 \times 5$;
- The number of neurons of the second hidden recursive layer is $96 \times 3$;
- The number of neurons of the output layer is $96$;
- The optimizer is "Adam", with $40$ epochs.

## 3  Results and Conclusions

In Figure 2, we present the load predictions for the community of buildings during the beginning of March 2016, after training the models for 2 consecutive months (Jan and Feb 2016). The above graphs illustrate well the performance of the predictive models and allow us to visually assess how correctly the AI models have been trained to learn the patterns of the load profile.

The Facebook`s Prophet model provides Human-in-the-Loop forecasts and gives the expert the ability to customize the model by applying their experience and external knowledge, e.g., in terms of calendar day labelling, local holiday effects, additive conditions, and annual trends, making it a sophisticated but potentially more accurate forecasting tool. Prophet provides the smoothest forecast profiles as compared to the rest of the models, while it accurately predicts the 8am and 6pm peaks in the community load consumption.

On the contrary, the LSTM forecast profiles try to capture the peaks in the load consumption and are less smooth compared to the Prophet's forecast profiles. Also, it was found that when the number of training days increases then the variance of the predicted values, relative to some average baseline profile, increases. Despite this, the network perfectly captures peak consumption values during the morning and evening, which is an undoubted advantage compared to the other models.

To test the accuracy of the forecasts, all models were simulated using data from 2016, i.e., 365 days. The simulations were conducted in the day-ahead forecast mode in the cross-validation way, which is close to the real situation of using the models. Based on the results of each day, the RMSE was calculated, averaged over the analyzed time interval. Since the cycle is moving step by step forward, the errors are obtained from a running average RMSE, Figure 3.

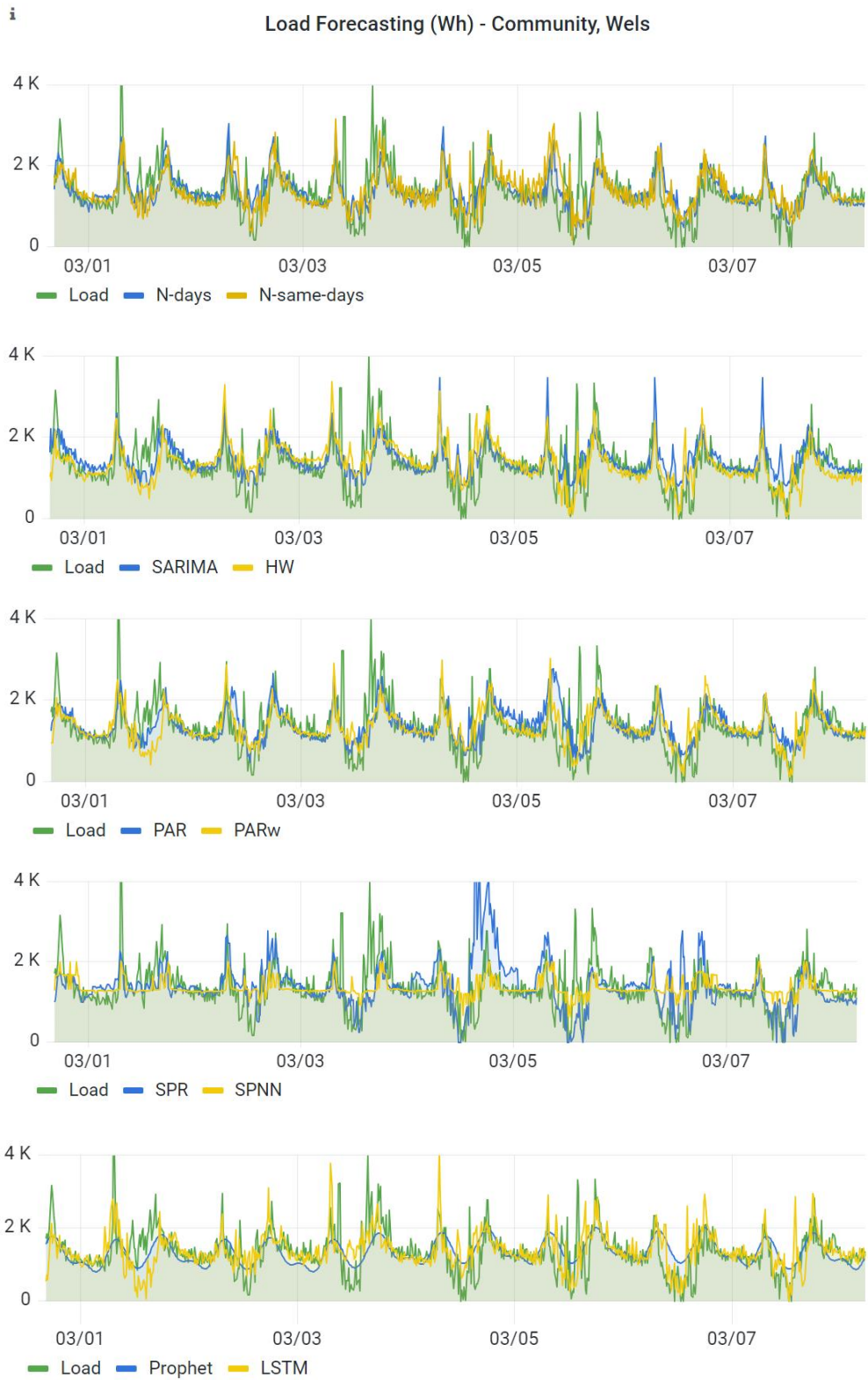

*Figure 2 Load predictions for residential buildings community (Wels, Upper Austria / March 2016)*

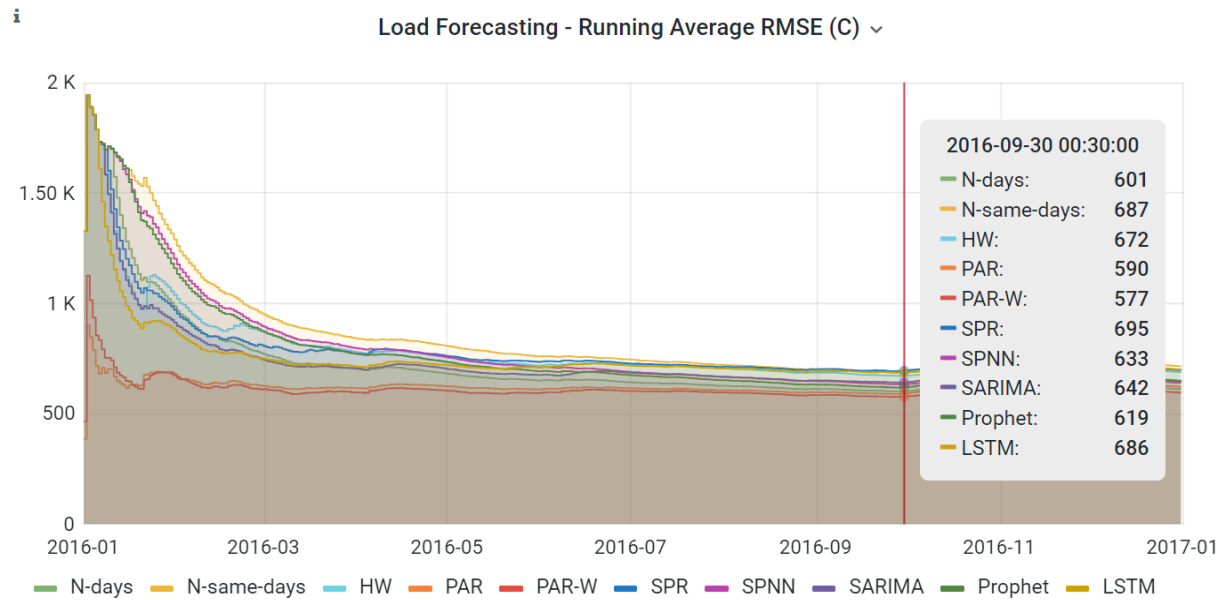

*Figure 3 Accuracy estimation of residential buildings community predictions via running average RMSE (Wels, Upper Austria / 2016)*

In order to provide an accuracy assessment that is not tied to the specific load amplitude values of the target community, the relative error was instead calculated. This is defined as the final value of the running average RMSE of the year divided by the mean community load. The overall accuracy of the models is also depicted in Table 1, which is calculated over 2016.

*Table 1 Comparison of AI-powered electricity load prediction methods with respect to relative average RMSE for the community of buildings.*

| Prophet | LSTM | N-same-days | N-days | HW | SARIMA | PAR | PAR-W | SPR | SPNN |
|---|---|---|---|---|---|---|---|---|---|
| 0.667 | 0.653 | 0.731 | 0.621 | 0.678 | 0.634 | 0.543 | 0.532 | 0.682 | 0.681 |

In this table, we see a comparison of the relative average RMSE of several standard and modern AI-powered models (namely Prophet and LSTM). AI-powered black-box model approaches, such as Prophet and LSTM, and machine learning models such as HW and SARIMA are generic time-series forecasting approaches that require computationally intensive training with several months of historical data. Furthermore, such models also require a careful hyperparameter configuration. Instead, the PAR and N-days models, designed specifically for electricity load-forecasting, exceed or match the performance of black-box forecasting models while they are characterized by low computational and configuration complexity. Furthermore, seasonal persistence-based regressive models use domain knowledge, but they are limited in the amount of statistical information that can be extracted from the observation sequences of measurements. Finally, we also observe that incorporating the weather forecasts in the PAR-W model maintained or slightly improved the forecasting accuracy, however the improvement is rather limited. This should be attributed to the fact that the forecasting models without the weather data features are already adapting to the variations in the electricity load due to weather.